\documentclass[runningheads]{llncs}
\usepackage[T1]{fontenc}
\usepackage{graphicx}
\usepackage{booktabs}
\usepackage{amsfonts}
\usepackage{amssymb}
\usepackage{amsmath}
\usepackage{mathrsfs}
\usepackage{indentfirst}
\usepackage{multirow}
\usepackage{multicol}
\usepackage{float}
\usepackage{color}
\usepackage{wrapfig}
\usepackage{graphicx}
\usepackage{pifont}
\usepackage{color}
\usepackage{caption}
\usepackage[table]{xcolor}
\usepackage{array} 
\usepackage[misc]{ifsym}
\begin{document}
\title{Negative Prototypes Guided Contrastive Learning for Weakly Supervised Object Detection}
\toctitle{Negative Prototypes Guided Contrastive Learning for Weakly Supervised Object Detection}
\titlerunning{Negative Prototypes Guided Contrastive Learning for WSOD}
%

\author{Yu Zhang \and
Chuang Zhu(\Letter) \and
Guoqing Yang \and Siqi Chen}

\tocauthor{Yu Zhang,Chuang Zhu,Guoqing Yang,Siqi Chen}
\authorrunning{Y. Zhang et al.}
%
\institute{School of Artificial Intelligence, Beijing University of Posts and Telecommunications, Beijing, China }
\maketitle              
\begin{abstract}
Weakly Supervised Object Detection (WSOD) with only image-level annotation has recently attracted wide attention. Many existing methods ignore the inter-image relationship of instances which share similar characteristics while can certainly be determined not to belong to the same category. Therefore, in order to make full use of the weak label, we propose the Negative Prototypes Guided Contrastive learning (NPGC) architecture. Firstly, we define Negative Prototype as the proposal with the highest confidence score misclassified for the category that does not appear in the label. Unlike other methods that only utilize category positive feature, we construct an online updated global feature bank to store both positive prototypes and negative prototypes. Meanwhile, we propose a pseudo label sampling module to mine reliable instances and discard the easily misclassified instances based on the feature similarity with corresponding prototypes in global feature bank. Finally, we follow the contrastive learning paradigm to optimize the proposal's feature representation by attracting same class samples closer and pushing different class samples away in the embedding space. Extensive experiments have been conducted on VOC07, VOC12 datasets, which shows that our proposed method achieves the state-of-the-art performance.

\keywords{Weakly supervised learning \and Object detection \and Contrastive learning.}
\end{abstract}

\section{Introduction}
Object detection is a classic computer vision task that jointly estimates class labels and bounding boxes of individual objects. In the last few decades, supervised learning of object detection has achieved remarkable progress thanks to the advances of Convolutional Neural Networks (CNNs) \cite{Girshick2014,Girshick2015,Ren2017}. However, the supervision of object detection training process often requires precise bounding boxes labels at a large scale, which is very labor-intensive and time-consuming.

Weakly supervised object detection (WSOD) \cite{Bilen2016} has recently attracted wide attention due to its greatly substitution of only image-level annotated datasets for precise annotated datasets in training process. Most existing methods are based on Multiple Instance Learning (MIL) \cite{Bilen2016,Tang2018,Tang2017,Ren2020,Huang2020} to transform WSOD into a multi-label classification task. \cite{Tang2017} tended to select the proposal with high confidence as the pseudo label and adopted multiple branch to refine the original proposal to gain more precise bounding-box, which has become the pipeline for numerous subsequent studies. 

However, with only image-level supervision, the classifier always faces the problem of \textbf{instance ambiguity} and \textbf{partial detection}. Instance ambiguity refers to the tendency to have missing instances or multiple grouped instances. Partial detection means that the detector tends to detect the most discriminative part of the target objects, which is also an inherent defect of the CNN network \cite{Gao2021}. Thus, there is still a large performance gap between weakly (mAP$=$56.8\% in VOC07) \cite{Huang2020} and fully (mAP$=$89.3\% in VOC07) \cite{Ghiasi2020} supervised object detectors. Different methods \cite{Wan2019,Huang2020,Kosugi2019,Ren2020,Shen2020,Zhong2020,Zhang2022,Lv2022,Dong2021,Sui2022,Seo2022} have been introduced to mitigate the above mentioned problems of WSOD. However, these methods generally lack full exploitation of the given limited annotation information. They mainly focus on the single input image itself, ignoring the corresponding relationship of instances in the whole dataset.

\begin{figure}[tp]
    \centering
    \includegraphics[scale=0.73]{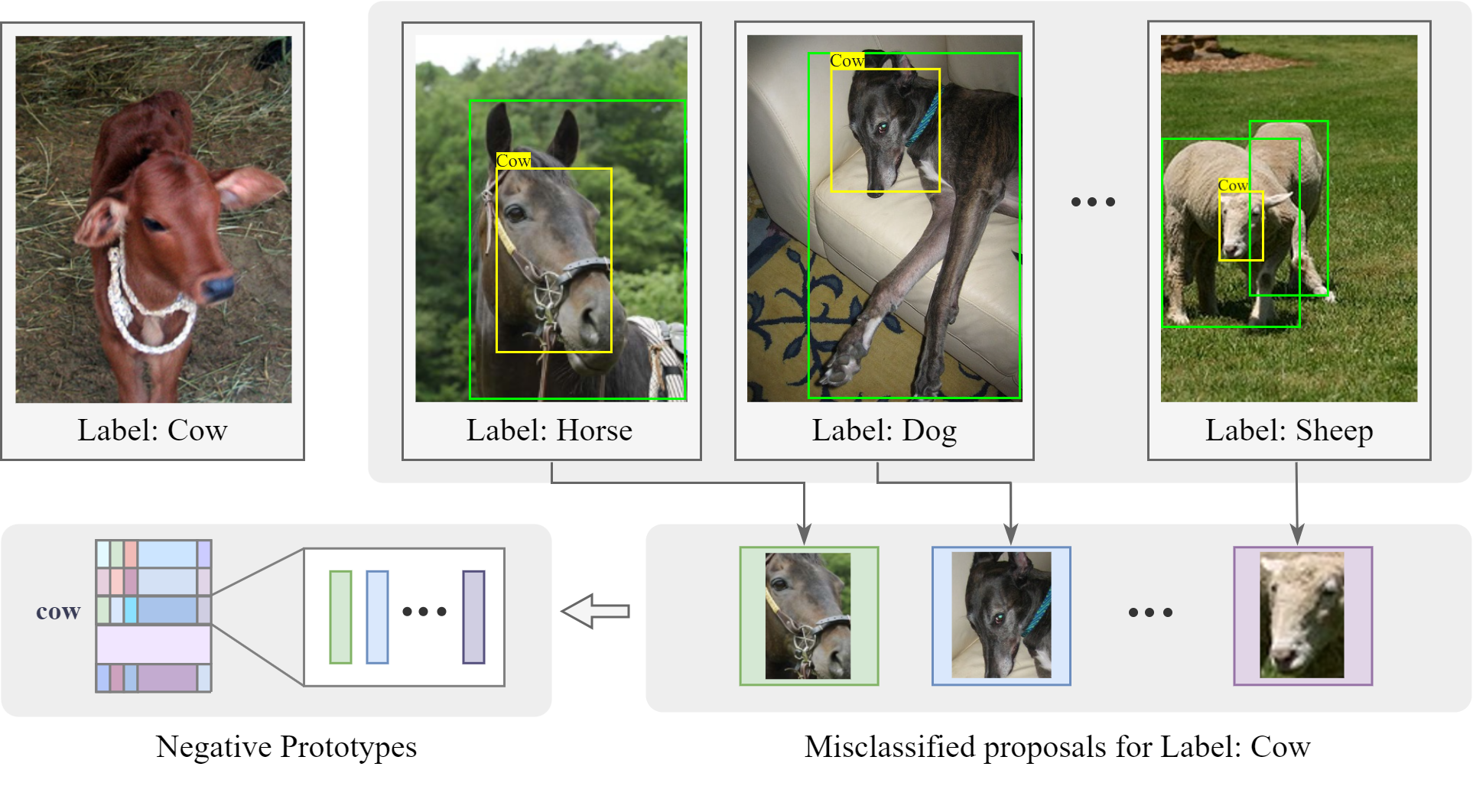}
    \caption{Illustration for negative prototypes. The green box in each of the three top-right images refers to the ground truth bounding box of the category ``Horse'', ``Dog'' and ``Sheep'', respectively. The yellow boxes refer to the misclassified proposals for the category ``Cow''. It is clear that ``Cow'' does not appear in any of the three images, while there are still proposals mistakenly detected as ``Cow''. We consider such proposals as negative prototypes for the category ``Cow''. We then extract the feature representations of these proposals and store them in the negative prototypes bank.}
    \label{fig:neg_show2}
\end{figure}


Therefore, we think of mining the hidden inter-image category information in the whole dataset. Instances belonging to the same category share similar characteristics, and we consider the typical features of the same category in the whole dataset as class positive prototypes. In contrast, we propose the concept of negative prototypes as the proposals with high confidence score misclassified for the category that does not appear in image label, which is illustrated in detail in Fig.~\ref{fig:neg_show2}. Observation reveals that negative prototypes always contain valuable category-specific discriminative features. The detector tends to produce false predictions in the regions containing category discriminative features due to the overfitting of discriminative regions (e.g. the heads of the dog and the horse). By leveraging the positive prototypes, we can retrieve several missing instances, and likewise, by leveraging the negative prototypes, we are also able to alleviate the problem of partial detection.

\begin{figure}[tp]
    \centering
    \includegraphics[scale=1]{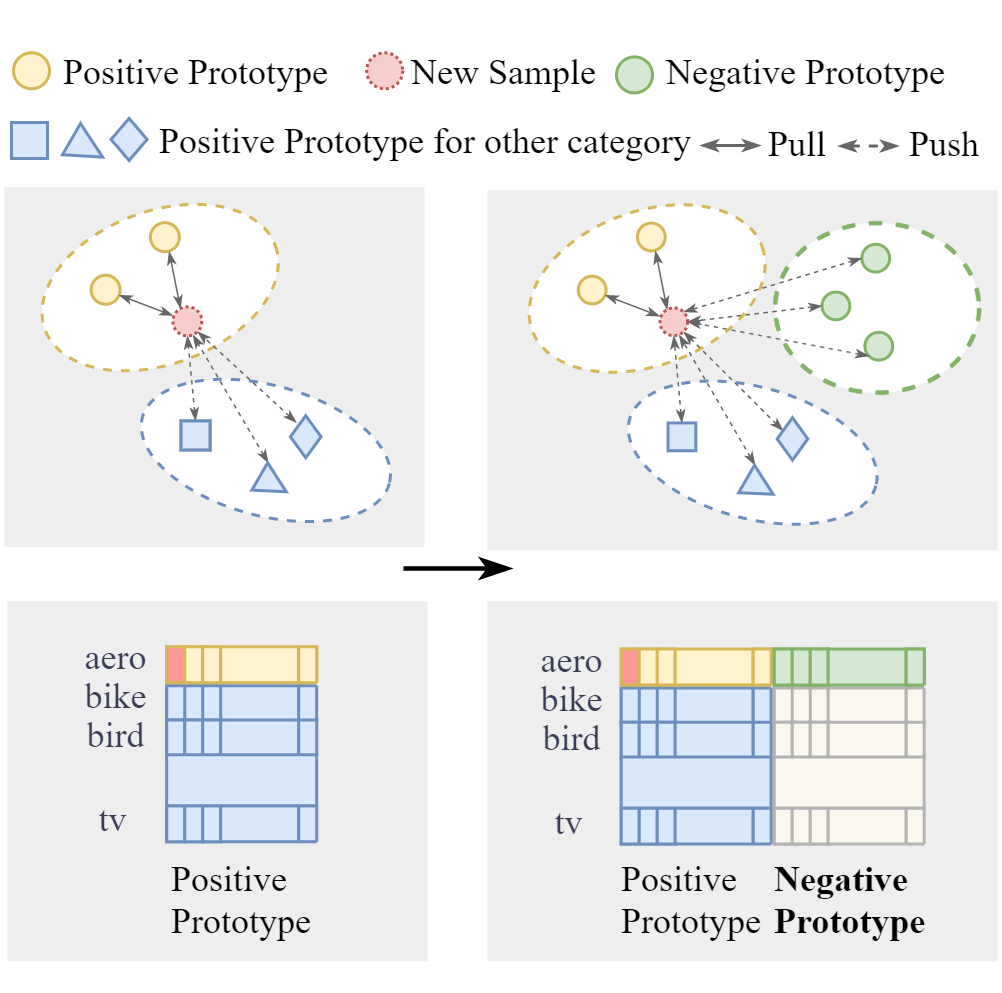}
    \caption{Comparison of classic contrastive learning (left) and our contrastive learning (right). We proposed the concept of Negative Prototypes (proposal mis-classified for category which has similar characteristics to current category while can certainly be determined not to belong to) and construct a global feature bank to store both positive prototype and negative prototype.}
    \label{fig:topimage2}
\end{figure}

In this paper, we propose a global negative prototypes guided contrastive learning weakly supervised object detection framework (NPGC). Our intuition is to fully exploit both visually correlated and visually discriminative category information in the whole dataset to improve the object classification ability of the weakly supervised detector. We construct an online updated global feature bank to store multiple class Positive Prototypes (PP) and Negative Prototypes (NP) from the whole dataset. Meanwhile, we design a novel Pseudo Label Sampling (PLS) module, which is used to mine the missing instances and punish overfitted instances that are prone to be partially detected. Based on the average feature similarity of candidate proposals and the positive prototypes of the same category, we can obtain a threshold $\tau_{pos}$ to mine proposals that might be omitted. Similarly, according to the average feature distance of candidate proposals and the negative prototypes of the same category with maximum similarity, a threshold $\tau_{neg}$ can also be obtained so as to discard the partial overfitted instances. Afterwards, as shown in Fig.~\ref{fig:topimage2}, we leverages a contrastive learning paradigm to narrow the distance in representation space between positive sample pairs and push the distance between negative sample pairs.

Our key contributions can be summarized as follows:
\begin{itemize}
\item[$\bullet$] First, we construct an elaborate global negative prototypes guided contrastive learning weakly supervised object detection framework. 
\end{itemize}

\begin{itemize}
\item[$\bullet$] Second, we propose negative prototypes which contain valuable category-specific discriminative features. We construct an online updated global feature bank to store both class positive prototypes and negative prototypes, and then leverage contrast learning loss to optimize it.
\end{itemize}

\begin{itemize}
\item[$\bullet$] Third, we devise a pseudo label sampling module, which utilized inter-image information from the global feature bank into pseudo proposal selection. This module effectively enables detector to mine the missing instances and simultaneously punish overfitted instances to alleviate the discriminal part detection problem.
\end{itemize}


\section{Related Work}
\label{sec:relatedwork}
\subsection{Weakly Supervised Object Detection}
Bilen \cite{Bilen2016} unifies deep convolutional network and Multi-Instance Learning (MIL) in an end-to-end WSOD network called Weakly Supervised Deep Detection Network (WSDDN) for the first time. As an improvement to WSDDN, Tang \textit{et al.}~\cite{Tang2017} gradually optimizes the predict bounding boxes by selecting high confidence region as pseudo label and adding an Online Instance Classifier Refinement module (OICR). Based on \cite{Tang2017}, in order to further improve detector's performance, Tang \textit{et al.}~\cite{Tang2018} introduces a Proposal Clustering Learning (PCL) method for candidate proposals, so that proposals with similar features could be clustered together as much as possible. More recently, Huang \textit{et al.}~\cite{Huang2020} proposes Comprehensive Attention Self-Distillation (CASD) framework that aggregate attention maps of input-wise and layer-wise to reach more balanced feature learning. Yin \textit{et al.}~\cite{Yin2021} devises an Instance Mining framework with Class Feature Bank (IM-CFB), which uses the uses the top-similarity scored instance to improve proposal selection. Seo \textit{et al.}~\cite{Seo2022} proposes a minibatch-level instance labeling and Weakly Supervised Contrastive Learning (WSCL) method with feature bank, while it may encounter the situation that the same category does not appear in the same minibatch. Inspired by \cite{Seo2022}, we propose a global class feature bank strategy and innovatively merge the prototypes of category negative samples, instead of solely employing the positive prototypes.

\subsection{Contrastive Learning}
Recently, there has been a trend towards exploring contrastive loss for representation learning. The idea of contrastive learning is to pull the samples from the positive pair closer together and push the samples from the negative pair apart. For instance, Hjelm \textit{et al.}~\cite{hjelm2018learning} propose Deep InfoMax to maximize the mutual information between the input and output of a deep network for unsupervised representation learning. More recently, Chen \textit{et al.}~\cite{Chen2019} presents a method for learning visual representations, which maximizes the agreement between different augmented views of the same image via a contrastive loss. He \textit{et al.}~\cite{He} proposes Momentum Contrast (MoCo), which utilizes a memory bank to store instance features. The purpose is to learn the representation by matching features of the same instance in different augmented views.  Tian \textit{et al.}~\cite{tian2020contrastive} extends the input to more than two views. These methods are all based on a similar contrastive loss associated with Noise Contrastive Estimation (NCE) \cite{Gutmann2009}. Oord \textit{et al.}~\cite{oord2018representation} proposed Contrastive Predictive Coding (CPC) that learns representations for sequential data. We choose the InfoNCE loss from \cite{oord2018representation} to minimize the distance between samples of the same category and maximize the distance between samples of different categories.

\section{Proposed Method}
\begin{figure*}[tp]
    \centering
    \includegraphics[scale=0.5]{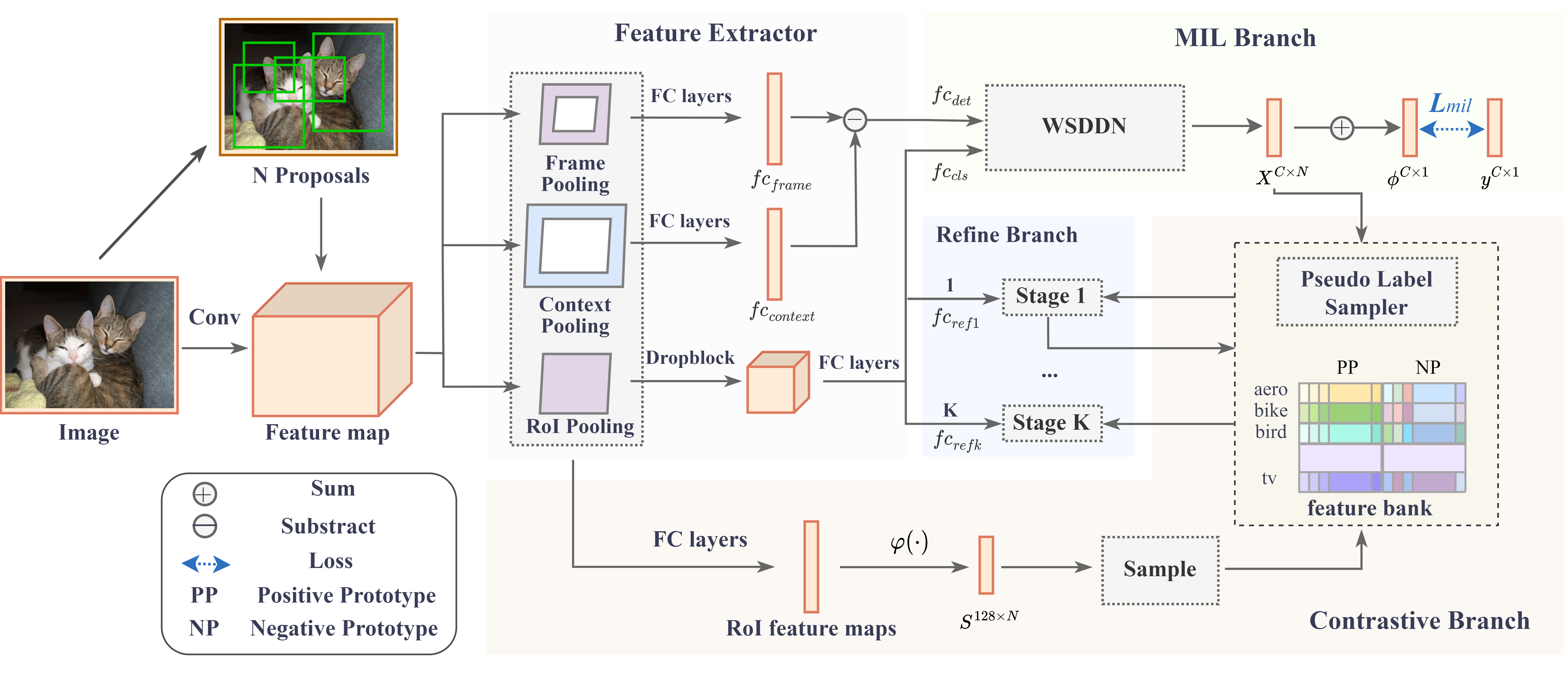}
    \caption{Overall architecture of the proposed method. NPGC consists of four major components: Feature extractor, MIL branch, Contrastive branch, and Online instance refine branch. We constructed a global feature bank to store both positive prototypes and negative prototypes, which utilized contrastive learning to pull close the samples from the positive pair and to push apart the samples from the negative pair. We employ a pseudo label sampling module to mine the missing instances and punish overfitted instances.}
    \label{fig:neg2}
\end{figure*}

In this paper, we introduce a negative prototypes guided contrastive learning weakly supervised object detection framework.
The overall architecture of the proposed network is shown in Fig~\ref{fig:neg2}.
We employ a MIL branch and an instance refinement branch as the basic network. On this basis we utilised a context-based feature extraction module to obtain more effective feature representation and designed a novel contrastive branch to employ the hidden inter-image information. 

\subsection{Preliminaries}

Formally, given a weakly supervised dataset $D$, we denote $ I \in \mathbb{R}^{h\times w \times3} $ as an input image from $D$. The image-level category label $y = \{ y_{1},\ldots,y_{C} \} \in \mathbb{R}^{C \times 1} $, where $C$ is the number of weakly supervised dataset categories. The corresponding region proposals pre-generated are $ R = \{ r_{1}, \ldots, r_{N} \}$, where $N$ is total number of proposals.

\subsubsection{MIL Branch}
For an input image $I$ and its region proposals $R$, a CNN backbone firstly extracts the image feature map $F$. $F$ is then fed into the feature extractor module containing different pooling layers and two Fully-Connected (FC) layers to obtain proposal feature vectors $fc_{cls}$ and $fc_{det}$. Subsequently, proposal features $fc_{cls}$ and $fc_{det}$ pass through MIL branch according to WSDDN \cite{Bilen2016}, which includes two streams to produce classification score matrices $X_{cls} \in \mathbb{R}^{C \times N}$ and detection score matrices $X_{det} \in \mathbb{R}^{C \times N}$, respectively. $X_{cls}$ normalized by a softmax layer $\sigma(\cdot)$ along the classes (rows) representing the probability of a region $r$ being classified as category $c$, whereas $X_{det}$ computed along the regions (columns) representing the probability of whether detecting region $r$ for category $c$ are obtained.
The final proposal score $X \in \mathbb{R}^{C \times N}$ is computed via an element-wise product $X=\sigma(X_{cls}) \odot \sigma(X_{det})$. The image score $\phi \in \mathbb{R}^{C \times 1}$ is obtained by the sum over all proposals, $\phi = \sum_{r=1}^N X_{r} $, and the following multi class cross entropy is minimized,
\begin{equation}
L_{mil}=-\sum_{c=1}^C \{ y_{c}\log \phi_{c}+(1-y_{c})\log (1- \phi_{c}) \} \label{eq:lmil}
\end{equation}

where $\phi_{c}$ equals to image score $\phi$ for the $c$-th class, $y_{c}$ represents whether an object of category $c$ is presented in the image.

\subsubsection{Online Instance Refine Branch}
For the $k$-th refine branch ($k\in\{1,\ldots,K\}$), $fc_{ref^{k}} \in \mathbb{R}^{(C+1) \times N}$ is the input proposal feature vector (The (C+1)-th category refers to background class). Each refinement stage is supervised by the previous stage, thus the pseudo ground truth label $\hat{y}_{c,r}^{k} \in \mathbb{R}^{(C+1) \times N}$ for stage $k$ is generated from the last stage’s output. 
Following the general pipeline \cite{Yang2019} an extra regression stream is added to regress bounding boxes online.
Overall, the instance refinement loss $L_{ref}^{k}$ is defined as,
\begin{equation}
L_{ref}^{k}=-\{ \frac{1}{N^k}\sum_{r=1}^{N^k}\sum_{c=1}^{C+1}\hat{y}_{c,r}^{k}\log x_{c,r}^{k}-\frac{1}{G^k}\sum_{r=1}^{G^k}Smooth_{L1}(t_{r}^{k},\hat{t_{r}^{k}})\} \label{eq:lref}
\end{equation}

where $N^k$ is the number of proposals and $G^k$ is the total number of positive proposals in the $k$-th branch. $t_{r}$ and $\hat{t_{r}}$ are the coordinate offsets and sizes of the $r$-th predicted and ground truth bounding-box.

\subsection{Feature Extractor}
Inspired by \cite{Vadim2016}, we extract three different features to represent each object proposal, which are the RoI feature, the context feature, and the frame feature, respectively. Specifically, the RoI feature is to represent the content of each proposal. The context feature is to represent the outer context content of each proposal, while the frame feature is to represent the inner context content of each proposal.

To represent the location of each proposal, we follow \cite{Vadim2016,Zhang2022} to subtract the pooled context feature $fc_{context}$ from the frame feature $fc_{frame}$ to obtain the input representation of the detection branch $fc_{det}$. Meanwhile, we leverage a dropblock to randomly mask out some blocks of the RoI feature map. We then let it go through two fc layers and serve as the input representation of the classification $fc_{cls}$ and refine branch $fc_{ref^k}, k \in \{1,\cdots,K\}$. By considering more information of the surrounding parts of the proposal, the extracted feature contains more location information and can effectively alleviate the problem of partial detection.
\subsection{Contrastive Branch}
For each image, all we know is its corresponding image level label, thus we intend to make as much use of this information as possible. Instances of the same category share similar characteristics, thus it is possible for us to extract several positive instance prototypes of each category from the entire data set, which is useful for alleviating the problem of missing instances.  At the same time, we can also extract the mis-classified instances corresponding to certain category from the whole data set and treat them as negative instance prototypes, which share the similar discriminative features of the highly overfitted region. We can leverage this proporty to mitigate the problem of partial detection.

Following \cite{Seo2022}, we construct a similarity head $\varphi(\cdot)$ as shown in Fig~\ref{fig:neg2}, which maps the input RoI feature vectors to $S \in \mathbb{R}^{128 \times N}$ a 128-dimensional embedding space. For each ground truth category $c$ from image $I$, we choose the top ranking proposal $\tilde{r}_{c,m} = \arg\max \limits_{N}(X_{c})$ from the final proposal score $X \in \mathbb{R}^{C \times N}$, where $m$ is the proposal index. And $\tilde{s}_{c,m}$ denoted as the corresponding feature representation of the top ranking proposal in the embedding space. And we store it into the positive feature bank. 

\subsubsection{Negative Prototypes} 
Our objective is to discover the negative prototypes of each category, so as to drive the detector's predictions away from the corresponding negative prototypes.

It can be calculated from the final proposal score matrix $X \in \mathbb{R}^{C \times N}$ that each image, except for categories in its ground truth label, gives out the index of the highest confidence score proposal mis-classified for other categories. This instance exactly is the negative prototype of its predicted category. Since we know explicitly that this proposal's prediction category is by no means exist in this image, yet it appears to be this category with high confidence. 
For example, in an input image labeled with ``cat”, one of the proposals is predicted to be ``dog'' with high confidence, but obviously ``dog'' shouldn't have been existed in this image. Therefore, we can conclude that this proposal has feature expression which tends to be mis-classified with ``dog'' while it is certainly not ``dog''. In this case, we suppose such proposal a negative prototype of category ``dog'', then we take its representation feature and store it into the negative feature bank.

The total global feature bank is denoted as $M= \bigcup_{c=1}^{C} S_{c}^{pos} \cup S_{c}^{neg}$, where $S^{pos}$ is the positive prototype bank and $S^{neg}$ is the negative prototype bank. 
For each selected negative prototype representation $s_{c,i}$, we select the most similar feature $s_{c,j}^{neg}$ from $S_{c}^{neg}$ to maximize the assistance of the current instance.
\begin{equation}
s_{c,j}^{neg}=r\cdot s_{c,j}^{neg}+(1-r)\cdot s_{c,i}\label{eq:update}
\end{equation}

where $r$ is the momentum coefficient \cite{He}, $s_{c,i}$ refers to the newly selected negative prototype, $s_{c,j}^{neg}$ refers to the the most similar feature with $s_{c,i}$ from $S_{c}^{neg}$. And the bank updating strategy is the same for the positive prototype bank.

\subsubsection{Pseudo Label Sampling Module} We construct a pseudo label sampling module to mine the missing instances and discard the overfitted instances. 

We first calculate the representation feature similarity $sim(\cdot)$ between the top ranking proposal $\tilde{s}_{c,m}$ and the positive prototypes $s_{c}^{pos}$ from the positive prototype bank. The average similarity is regarded as the threshold $\tau_{pos}$ for mining positive samples.

\begin{equation}
\tau_{pos}=\frac{1}{\left\vert S_{c}^{pos} \right\vert} \sum_{i=1}^{\left\vert S_{c}^{pos} \right\vert} sim(\tilde{s}_{c,m},s_{c,i}^{pos})\label{eq:tpos}
\end{equation}

For each candidate proposal $r \in R$, we caculate the similarity between $s_r$ and the top ranking proposal feature $\tilde{s}_{c,m}$, from which we can mine proposals might be omitted by selecting candidate proposals whose similarity exceed $\tau_{pos}$. 
\begin{equation}
sim(s_{r},\tilde{s}_{c,m})>\tau_{pos}\label{eq:simpos}
\end{equation}

Accordingly, we calculate the similarity between each candidate proposal feature $s_r$ and its corresponding negative prototype $\tilde{s}_{c,r}^{neg}$ with maximum similarity, where $\tilde{s}_{c,r}^{neg}=\arg\max \limits_{i}( sim(s_{r},s_{c,i}^{neg})), i=\{1,\cdots,\left\vert S_{c}^{neg} \right\vert\}$. The average similarity is regarded as the threshold $\tau_{neg}$ for discarding negative samples
\begin{equation}
\tau_{neg}=\frac{1}{\left\vert R \right\vert} \sum_{r=1}^{\left\vert R \right\vert} sim(s_{r},\tilde{s}_{c,r}^{neg})\label{eq:tneg}
\end{equation}

The feature similarity between current instance and its negative prototype represents the probability of belonging to easily mis-classified discriminal regions. Instances with low feature similarity , i.e. those below the threshold $\tau_{neg}$, means that the instance is most likely overfitted and should be discarded.
\begin{equation}
sim(s_{r},\tilde{s}_{c,r}^{neg})<\tau_{neg}\label{eq:simneg}
\end{equation}

\textbf{Contrastive Learning:} We use contrastive learning to optimize the feature representations for the proposals by attracting positive samples closer together and repelling negative samples away from positives samples in the embedding space. To obtain more views of samples for contrastive learning, we apply the same feature augmentation methods following \cite{Seo2022}.
\begin{equation}
L_{cont}=-\frac{1}{\left\vert M \right\vert}\sum_{i=1}^{\left\vert M \right\vert} \log \frac{exp(s_{i}\cdot s_{+}/\varepsilon)}{exp(s_{i}\cdot s_{+}/\varepsilon)+\sum_{S_{-}}exp(s_{i}\cdot s_{+}/\varepsilon)}\label{eq:lcont}
\end{equation}

where $M= \bigcup_{c=1}^{C} S_{c}^{pos} \cup S_{c}^{neg}$, and $\varepsilon$ is a temperature parameter introduced in \cite{NEURIPS2020_d89a66c7}.
We use the contrastive loss to pull $s_{i}$ close to $s_{+}$ of the same class while pushing it away from $s_{-}$ both positive prototypes from other classes and its negative prototypes, and thus enhance the discrimination and generalization of current instance representation

Finally, the total loss of training the network is the combination of all the loss functions mentioned before.
\begin{equation}
L_{total}=L_{mil}+\sum_{k=1}^{K} L_{ref}^{k}+\lambda L_{cont}\label{eq:totalloss}
\end{equation}

\section{Experimental Results}
\subsection{Datasets} 

We evaluate our proposed method on both Pascal VOC 2007 and Pascal VOC 2012 \cite{article} , which are commonly used to assess WSOD performance. For the VOC datasets, we employ the trainval set (5,011 images in VOC 2007, 11,540 images in VOC 2012) for training and evaluate the model's performance on the test set. VOC 2007 and 2012 both contain 20 categories. We apply Mean average precision (mAP) with standard IoU threshold (0.5) to evaluate the object detection accuracy on the testing set. 

\subsection{Implementation Details}

We adopt the Imagenet \cite{5206848} pretrained model VGG16 \cite{Simonyan2014Very} as the backbone. For VGG16, following the previous methods \cite{Seo2022}, we replace a global average pooling layer with a RoI pooling layer, and remove the last FC layer leaving two FC layers, which all the heads including the similarity head are attached to. Proposal generating method such as Selective Search \cite{Uijlings2013SelectiveSF} and MCG \cite{inproceedings} are used for VOC dataset to generate initial proposals, and we use around 2,000 proposals per image. Then, the whole model is trained on 4 NVIDIA GeForce GTX 3090 with 24 GB GPU memory using a SGD optimizer with an initial learning rate of 0.01, weight decay of 0.0001 and momentum of 0.9 are used to optimize the model.

\setlength{\intextsep}{5pt plus 2pt minus 2pt}
\begin{wraptable}{r}{0.6 \textwidth}
\caption{Comparison of the State-of-the-arts methods on VOC07 and VOC12 of mAP(\%).}
\renewcommand{\arraystretch}{1.1}
\begin{tabular}{ p{3.5cm}| p{50pt}<{\centering} | p{50pt}<{\centering} }
      \toprule[0.75pt]
      \textbf{Method} & \textbf{mAP(\%) VOC07} & \textbf{mAP(\%) VOC12}\\
      \midrule[0.3pt]
      \midrule[0.3pt]
      WSDDN \cite{Bilen2016}$_{~\textup{CVPR'16}}$ & 34.8 & - \\
      OICR \cite{Tang2017}$_{~\textup{CVPR'17}}$ & 41.2 & 37.9\\
      PCL \cite{Tang2018}$_{~\textup{TPAMI'18}}$ & 43.5 & 40.6\\
      C-WSL \cite{Gao2018}$_{~\textup{ECCV'18}}$ & 46.8 & 43.0\\
      C-MIL \cite{Wan2018}$_{~\textup{CVPR'18}}$ & 50.5 & 46.7\\
      C-MIDN \cite{9009008}$_{~\textup{ICCV'19}}$ & 52.6 & 50.2\\
      WSOD2 \cite{Zeng2019}$_{~\textup{ICCV'19}}$ & 53.6 & 47.2\\
      SLV \cite{Chen2020}$_{~\textup{CVPR'20}}$ & 53.5 & 49.2\\
      MIST \cite{Ren2020}$_{~\textup{CVPR'20}}$ & 54.9 & 52.1\\
      CASD \cite{Huang2020}$_{~\textup{NIPS'20}}$ & 56.8 & 53.6\\
      IM-CFB \cite{Yin2021}$_{~\textup{AAAI'21}}$  & 54.3 & 49.4\\
      CPE \cite{Lv2022}$_{~\textup{TIP'22}}$ & 55.9 & 54.3\\
      NDI \cite{wang2022absolute}$_{~\textup{IJCAI'22}}$ & 56.8 & 53.9\\
      \rowcolor{lightgray} 
      \textbf{Ours} & \textbf{57.7} & \textbf{54.3} \\
      \bottomrule[0.75pt]
\end{tabular}
\label{table:comparison}
\end{wraptable}

The overall iteration numbers are set to 35,000, 70,000 for VOC 2007, VOC 2012. Following the previous methods \cite{Ren2020,Tang2017,Huang2020}, the inputs are multi-scaled to \{480, 576, 688, 864, 1000, 1200\} for both training and inference time. The final predictions are made after applying NMS of which threshold is set to 0.4 for both datasets. In the refinement branch, we set the number of refinement stages $K=3$. The bank size $M$ is set to 6, which is experimentally illustrated in Table.~\ref{fb}. The hyperparameter $\varepsilon$ from eq.~\eqref{eq:lcont} is set to 0.2 following the experiments conducted in \cite{NEURIPS2020_d89a66c7,Chen2019}. And hyperparameter $\lambda$ from eq.~\eqref{eq:totalloss} is set to 0.03 as explained in Sec.4.5.

\setlength{\intextsep}{5pt plus 2pt minus 2pt}
\begin{table*}[h]
  \caption{Comparison with the state-of-the-arts methods in terms of Per-class AP results on VOC07.}\label{perap07}
\renewcommand{\arraystretch}{1.2}
  \begin{center}
    \scalebox{0.62}{
    \begin{tabular}{l|c|c|c|c|c|c|c|c|c|c|c|c|c|c|c|c|c|c|c|c|r} 
      \toprule[0.75pt]
      \textbf{Method} & Aero & Bike & Bird & Boat & Bottle & Bus & Car & Cat & Chair & Cow & Table & Dog & Horse & Motor & Person & Plant & Sheep & Sofa & Train & TV & \textbf{mAP}\\
      \midrule[0.5pt]
      \midrule[0.5pt]
      WSDDN & 39.4 & 50.1 & 31.5 & 16.3 & 12.6 & 64.5 & 42.8 & 42.6 & 10.1 & 35.7 & 24.9 & 38.2 & 34.4 & 56.6 & 9.4 & 14.7 & 30.2 & 40.7 & 54.7 & 46.9 & 34.8 \\
      OICR & 58.0 & 62.4 & 31.1 & 19.4 & 13.0 & 65.1 & 62.2 & 28.4 & 24.8 & 44.7 & 30.6 & 25.3 & 37.8 & 65.5 & 15.7 & 24.1 & 41.7 & 46.9 & 64.3 & 62.6 & 41.2 \\
      PCL & 54.4 & 69.0 & 39.3 & 19.2 & 15.7 & 62.9 & 64.4 & 30.0 & 25.1 & 52.5 & 44.4 & 19.6 & 39.3 & 67.7 & 17.8 & 22.9 & 46.6 & 57.5 & 58.6 & 63.0 & 43.5\\
      C-WSL & 62.9 & 64.8 & 39.8 & 28.1 & 16.4 & 69.5 & 68.2 & 47.0 & 27.9 & 55.8 & 43.7 & 31.2 & 43.8 & 65.0 & 10.9 & 26.1 & 52.7 & 55.3 & 60.2 & 66.6 & 46.8\\
      C-MIL & 62.5 & 58.4 & 49.5 & 32.1 & 19.8 & 70.5 & 66.1 & 63.4 & 20.0 & 60.5 & \textbf{52.9} & 53.5 & 57.4 & 68.9 & 8.4 & 24.6 & 51.8 & 58.7 & 66.7 & 63.5 & 50.5\\
      C-MIDN & 53.3 & 71.5 & 49.8 & 26.1 & 20.3 & 70.3 & 69.9 & 68.3 & 28.7 & 65.3 & 45.1 & 64.6 & 58.0 & 71.2 & 20.0 & 27.5 & 54.9 & 54.9 & 69.4 & 63.5 & 52.6\\
      WSOD2 & 65.1 & 64.8 & 57.2 & \textbf{39.2} & 24.3 & 69.8 & 66.2 & 61.0 & 29.8 & 64.6 & 42.5 & 60.1 & 71.2 & 70.7 & 21.9 & 28.1 & 58.6 & 59.7 & 52.2 & 64.8 & 53.6\\
      SLV & 65.6 & 71.4 & 49.0 & 37.1 & 24.6 & 69.6 & 70.3 & 70.6 & 30.8 & 63.1 & 36.0 & 61.4 & 65.3 & 68.4 & 12.4 & \textbf{29.9} & 52.4 & 60.0 & 67.6 & 64.5 & 53.5\\
      MIST & 68.8 & \textbf{77.7} & 57.0 & 27.7 & 28.9 & 69.1 & 74.5 & 67.0 & 32.1 & \textbf{73.2} & 48.1 & 45.2 & 54.4 & 73.7 & 35.0 & 29.3 & \textbf{64.1} & 53.8 & 65.3 & 65.2 & 54.9\\
      CASD & - & - & - & - & - & - & - & - & - & - & - & - & - & - & - & - & - & - & - & - & 56.8\\
      IM-CFB & 64.1 & 74.6 & 44.7 & 29.4 & 26.9 & 73.3 & 72.0 & 71.2 & 28.1 & 66.7 & 48.1 & 63.8 & 55.5 & 68.3 & 17.8 & 27.7 & 54.4 & 62.7 & 70.5 & 66.6 & 54.3\\
      CPE & 62.4 & 76.4 & \textbf{59.7} & 33.8 & 28.7 & 71.7 & 66.1 & \textbf{72.2} & \textbf{33.9} & 67.7 & 47.6 & \textbf{67.2} & 60.0 & 71.7 & 18.1 & 29.9 & 53.8 & 58.9 & \textbf{74.3} & 64.1 & 55.9\\
      NDI & - & - & - & - & - & - & - & - & - & - & - & - & - & - & - & - & - & - & - & - & 56.8\\
    \rowcolor{lightgray} 
      \textbf{Ours} & \textbf{69.1} & 77.1 & 54.7 & 31.8 & \textbf{29.7} & \textbf{74.3} & \textbf{78.6} & 71.5 & 20.1 & 72.6 & 34.5 & 61.6 & \textbf{75.3} & \textbf{78.4} & \textbf{35.7} & 24.1 & 59.1 & \textbf{66.4} & 72.9 & \textbf{67.1} & \textbf{57.7}\\
      \bottomrule[0.75pt]
    \end{tabular}
    }
  \end{center}
\end{table*}

\subsection{Comparison with State-of-the-arts}

In Table~\ref{table:comparison}, we compare our proposed method with other state-of-the-art algorithms on PASCAL VOC07 and VOC12. Regardless of backbone structure, the results show that our method achieves the 57.7\% mAP and 54.3\% mAP in VOC07 and VOC12, respectively, which outperforms the other methods and reach the new state-of-the-art performance. 

\begin{figure*}[t]
    \centering
    \includegraphics[scale=0.7]{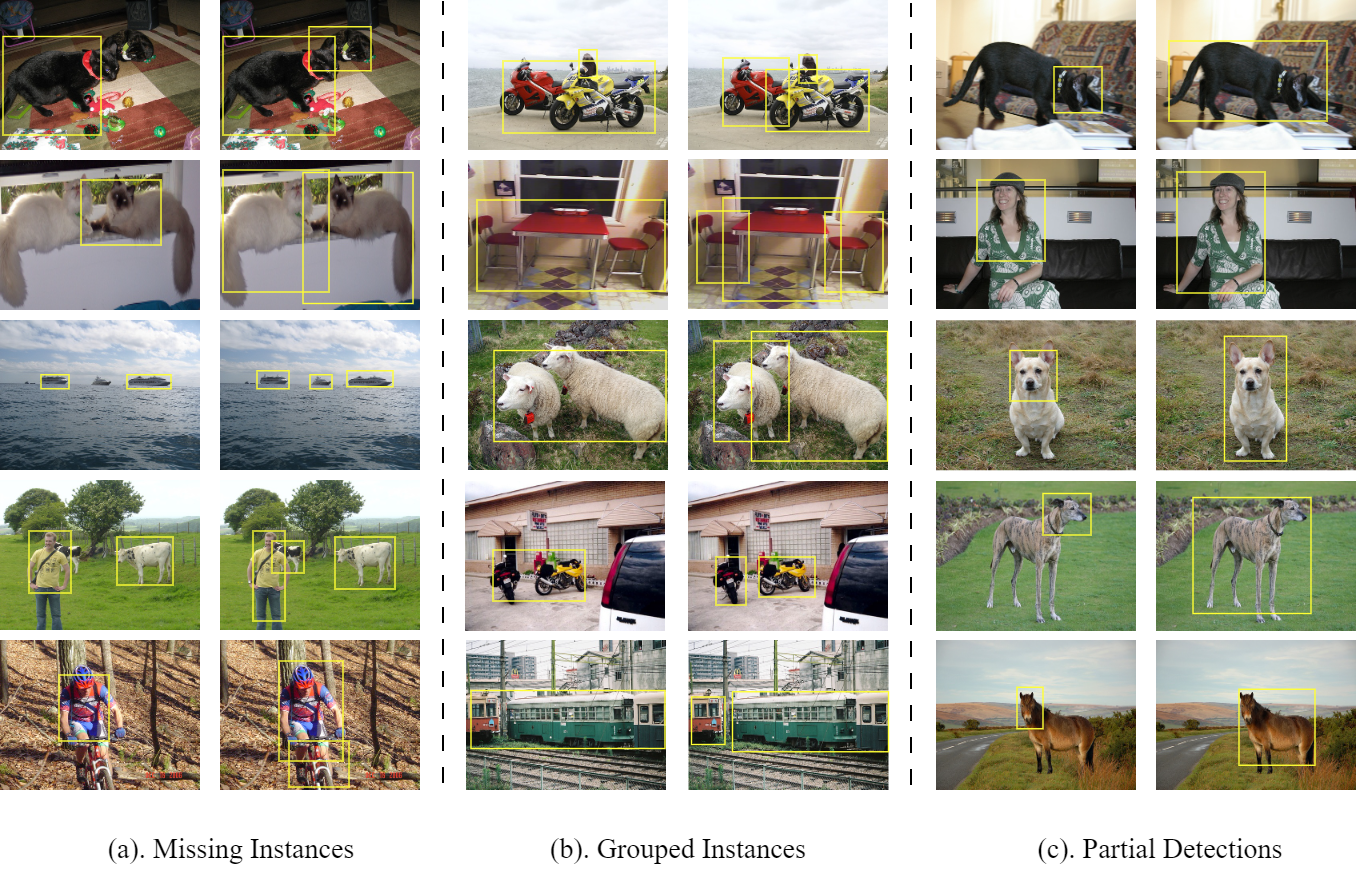}
    \caption{Qualitative results on VOC 2007 test set. The left columns show the results from OICR whereas the right columns show the results from our method.}
    \label{fig:compare}
\end{figure*}

It is shown in Fig.~\ref{fig:compare} that our method effectively addresses on the main challenges of WSOD compared to OICR \cite{Tang2017}.  The left columns show the results from OICR whereas the right columns show the results from our method. In (a) and (b), we investigate the effectiveness of our model in resolving the instance ambiguity problem which consists of missing instances and grouped instances, respectively. We can observe that 
many instances that have been ignored previously can be detected via our model. Meanwhile, in (b) we can also observe that grouped instances are separated into multiple bounding boxes. Moreover, the partial detection problem is largely alleviated shown in (c), especially for the categories with various poses such as dog, cat and person.

\begin{figure*}[th]
    \centering
    \includegraphics[scale=0.45]{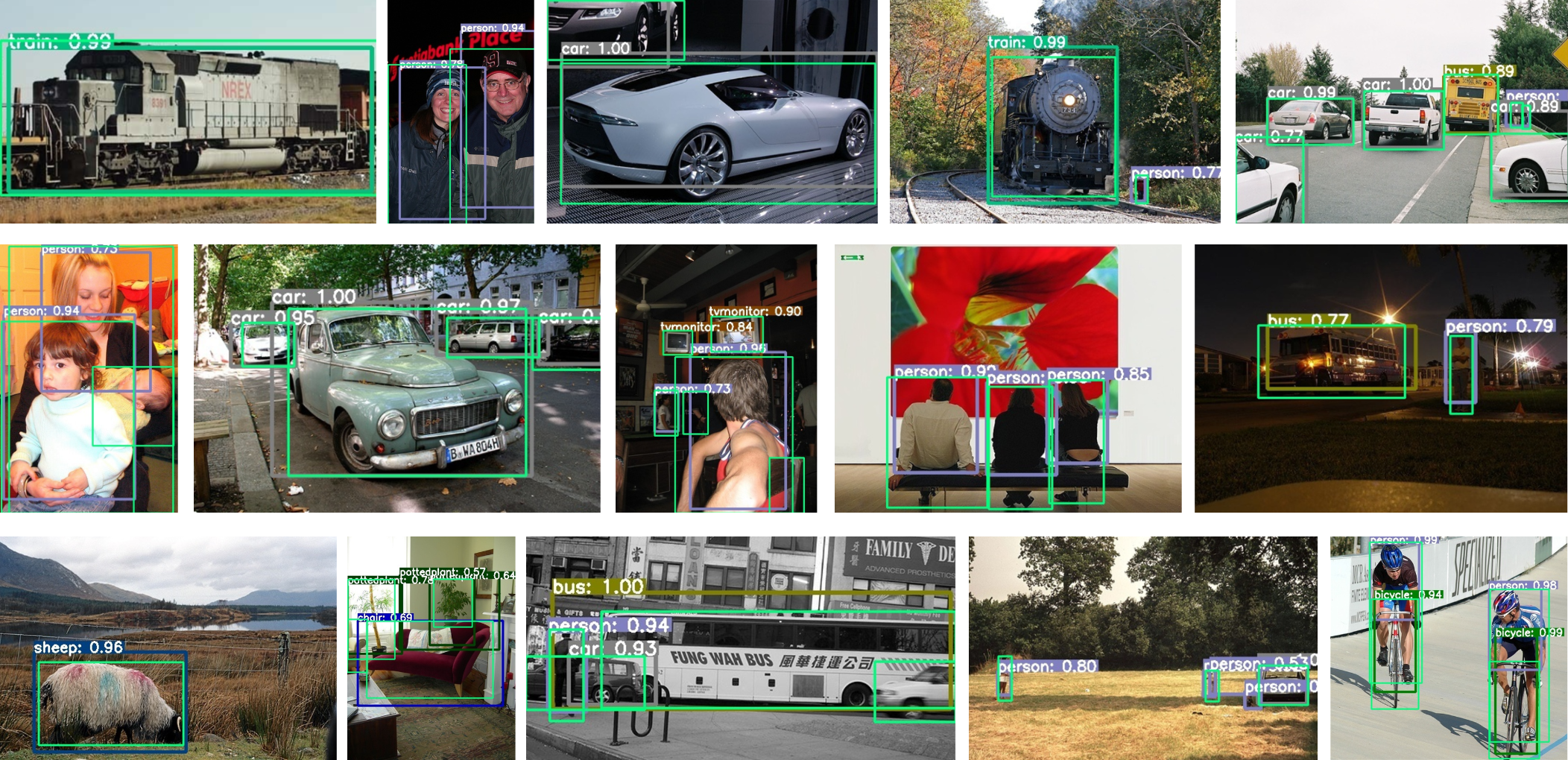}
    \caption{More detection results on VOC 2007 test set. Boxes in light green represent ground-truth boxes, and boxes in other colors represent the predicted bounding boxes and the confidence scores.}
    \label{fig:show}
\end{figure*}

More qualitative results are shown in Fig.~\ref{fig:show}, from which it can be seen that our model is able to mine the easily omitted multiple instances of the same category (car, person) and detect various objects of different classes (tvmoniter, pottleplant) in relatively complicated scenes.


\setlength{\intextsep}{5pt plus 2pt minus 2pt}
\begin{table*}[h]
  \caption{Comparison with the state-of-the-arts methods in terms of Per-class AP results on VOC12.}\label{perap12}
\renewcommand{\arraystretch}{1.2}
  \begin{center}
    \scalebox{0.62}{
    \begin{tabular}{l|c|c|c|c|c|c|c|c|c|c|c|c|c|c|c|c|c|c|c|c|r} 
      \toprule[0.75pt]
      \textbf{Method} & Aero & Bike & Bird & Boat & Bottle & Bus & Car & Cat & Chair & Cow & Table & Dog & Horse & Motor & Person & Plant & Sheep & Sofa & Train & TV & \textbf{mAP}\\
      \midrule[0.5pt]
      \midrule[0.5pt]
      OICR & 67.7 & 61.2 & 41.5 & 25.6 & 22.2 & 54.6 & 49.7 & 25.4 & 19.9 & 47.0 & 18.1 & 26.0 & 38.9 & 67.7 & 2.0 & 22.6 & 41.1 & 34.3 & 37.9 & 55.3 & 37.9 \\
      PCL & - & - & - & - & - & - & - & - & - & - & - & - & - & - & - & - & - & - & - & - & 40.6\\
      C-WSL & 74.0 & 67.3 & 45.6 & 29.2 & 26.8 & 62.5 & 54.8 & 21.5 & 22.6 & 50.6 & 24.7 & 25.6 & 57.4 & 71.0 & 2.4 & 22.8 & 44.5 & 44.2 & 45.2 & 66.9 & 43.0\\
      C-MIL & - & - & - & - & - & - & - & - & - & - & - & - & - & - & - & - & - & - & - & - & 46.7\\
      C-MIDN & 72.9 & 68.9 & 53.9 & 25.3 & 29.7 & 60.9 & 56.0 & \textbf{78.3} & 23.0 & 57.8 & \textbf{25.7} & \textbf{73.0} & 63.5 & 73.7 & 13.1 & 28.7 & 51.5 & 35.0 & 56.1 & 57.5 & 50.2\\
      WSOD2 & - & - & - & - & - & - & - & - & - & - & - & - & - & - & - & - & - & - & - & - & 47.2\\
      SLV & - & - & - & - & - & - & - & - & - & - & - & - & - & - & - & - & - & - & - & - & 49.2\\
      MIST & \textbf{78.3} & 73.9 & 56.5 & \textbf{30.4} & \textbf{37.4} & 64.2 & \textbf{59.3} & 60.3 & \textbf{26.6} & 66.8 & 25.0 & 55.0 & 61.8 & \textbf{79.3} & 14.5 & \textbf{30.3} & \textbf{61.5} & 40.7 & 56.4 & 63.5 & 52.1\\
      CASD & - & - & - & - & - & - & - & - & - & - & - & - & - & - & - & - & - & - & - & - & 53.6\\
      IM-CFB & - & - & - & - & - & - & - & - & - & - & - & - & - & - & - & - & - & - & - & - & 49.4\\
      CPE & - & - & - & - & - & - & - & - & - & - & - & - & - & - & - & - & - & - & - & - & 54.3\\
      NDI & - & - & - & - & - & - & - & - & - & - & - & - & - & - & - & - & - & - & - & - & 53.9\\
      \rowcolor{lightgray} 
      \textbf{Ours} & 75.4 & \textbf{75.3} & \textbf{59.1} & 29.6 & 30.6 & \textbf{69.9} & 56.8 & 63.0 & 23.3 & \textbf{71.3} & 25.3 & 63.1 & \textbf{66.4} & 76.7 & \textbf{19.0} & 25.5 & 61.4 & \textbf{56.7} & \textbf{66.6} & \textbf{70.5} & \textbf{54.3}\\
      \bottomrule[0.75pt]
    \end{tabular}
    }
  \end{center}
\end{table*}

\subsection{Ablation Study}
In this section, we make a comprehensive ablation study of the effect gains from different components, the sensitivity of hyperparameters, and the length of feature bank. The experiments are implemented on the VOC 2007 dataset. 

\subsubsection{Components effect}
We conduct experiments to prove the effectiveness of each component in our proposed method as shown in Table~\ref{hyper}, where PLS, CL, NP means the pseudo label sampling module, contrastive learning, negative prototypes mentioned in Sec 3.3, respectively. Our Baseline is the framework in Fig.~\ref{fig:neg2} without contrastive branch, which achieves 56.1\% mAP.

\begin{table}[h]
\caption{Ablation study on VOC 2007 dataset of different components in our method.}\label{tab1}
\setlength{\tabcolsep}{4pt}
\renewcommand{\arraystretch}{1.1}
  \begin{center}
    \begin{tabular}{c|c|c|c|c} 
      \toprule[0.75pt]
      \textbf{Baseline} & \textbf{PLS} & \textbf{CL} &
      \textbf{NP} &
      \textbf{mAP(\%)}\\
      \midrule[0.5pt]
      \midrule[0.5pt]
      $\surd$ &   &  & & 56.1    \\
      $\surd$ & $\surd$  & & & 56.8(+0.7) \\
      $\surd$ & $\surd$  &$\surd$ & & 57.2(+1.1) \\
      \rowcolor{lightgray}
      $\surd$ & $\surd$  &$\surd$ & $\surd$ & \textbf{57.7}(+1.6) \\
      \bottomrule[0.75pt]
    \end{tabular}
  \end{center}
\end{table}

We firstly analyze the effect of PLS and CL algorithm on our method NPGC. As shown in Table~\ref{tab1}, after applying PLS, our method achieves 56.8\% mAP with 0.7\% gains. After applyig both PLS and CL, it brings 1.1\% gains in mAP. Based on this, we append the nagative prototyes into former structure, and it reach 57.7\% mAP, which shows the effectiveness of our method. 

\setlength{\intextsep}{5pt plus 2pt minus 2pt}

\begin{table}[h]
\caption{Ablation study on different hyperparameters' value in our method.}\label{hyper}
\setlength{\tabcolsep}{4pt}
\renewcommand{\arraystretch}{1.1}
  \begin{center}
    \begin{tabular}{c||c|c|c|c|c} 
      \toprule[0.75pt]
      \textbf{$\lambda$} & 0.01 & 0.02 &
      0.03 &
      0.04 &
      0.05\\
      \midrule[0.5pt]
      \textbf{mAP(\%)} & 56.6 & 57.0 & \textbf{57.7} & 56.1 & 56.3   \\
      \bottomrule[0.75pt]
    \end{tabular}
  \end{center}
\end{table}

\subsubsection{Hyperparameters}
We provide the experiment results with different values of the hyperparameters we introduce. We conduct experiments on how to choose the loss parameter $\lambda$ from eq.~\eqref{eq:totalloss} in Table 4, and $\lambda=0.03$ achiveves the best result. In eq.~\eqref{eq:lcont}, we use the same values of $\varepsilon=0.2$ following the experiments conducted in other contrastive learning methods \cite{NEURIPS2020_d89a66c7,Chen2019}.

\begin{table}[h]
\caption{Ablation study on the length of feature bank in our method.}\label{fb}
\setlength{\tabcolsep}{4pt}
\renewcommand{\arraystretch}{1.1}
  \begin{center}
    \begin{tabular}{c||c|c|c|c} 
      \toprule[0.75pt]
      \textbf{$M$} & 2 & 4 &
      6 &
      8 \\
      \midrule[0.5pt]
      \textbf{mAP(\%)} & 55.6 & 57.0 & \textbf{57.7} & 56.4   \\
      \bottomrule[0.75pt]
    \end{tabular}
  \end{center}
\end{table}

\subsubsection{Length of feature bank}
We finally analyze the effect of the length of feature bank. If the length is too small, the feature bank is difficult to store the diversity of instance representations well, resulting in less kind of objects collected. And if the length is too large, it is easy to absorb some noisy information during the learning of instance representations and background proposals will be selected incorrectly.
In this paper, we recommend setting $M = 6$ to balance the number of stored instance features.

\section{Conclusion}

In conclusion, we presented a global negative prototypes guided contrastive learning weakly supervised object detection framework. We novelly introduce the concept of Negative Prototypes. Meanwhile, we construct a global feature bank to store both positive prototypes and negative prototypes, using contrastive learning to mine the hidden inter-image category information in the whole dataset.


\section{Ethical Statement}
This research was conducted in accordance with ethical guidelines and regulations. The paper aims to contribute to knowledge while upholding ethical standards.

\subsubsection{Acknowledgements} This work was supported by the National Key R\&D Program of China (2021ZD0109800), the National Natural Science Foundation of China (81972248) and BUPT innovation and entrepreneurship support program 2023-YC-A185.

%
%
%
\bibliographystyle{splncs04}
\bibliography{egbib}
%




\end{document}